\documentclass[conference]{IEEEtran}

\usepackage{psfrag,epsfig,graphics,url,amsmath,amssymb,fancyhdr,color}
\usepackage{special,times,graphicx}

\usepackage{helvet}  % DO NOT CHANGE THIS
\usepackage{courier}  % DO NOT CHANGE THIS
\usepackage{caption} % DO NOT CHANGE THIS AND DO NOT ADD ANY OPTIONS TO IT
\usepackage[ruled]{algorithm2e}
\usepackage{newfloat}
\usepackage{listings}
\usepackage[pdftex,hyperref,svgnames,table,xcdraw]{xcolor}

\usepackage[tight,footnotesize]{subfigure}
\usepackage{hhline}
\usepackage{booktabs}
\usepackage{multirow}
\usepackage[filecolor=black,citecolor=black,urlcolor=black,linkcolor=black,colorlinks,pdfpagelabels,pdfpagelayout=SinglePage]{hyperref}
\usepackage{breakurl}
\colorlet{todo}{FireBrick} % FireBrick prints better than red

\newcommand{\eg}{{\it e.g.}}

\newcommand{\etal}{{\it et al.}}
\newcommand{\comment}[1]{}

\newcommand{\fp}{\vspace*{0.03in}\noindent}

\renewcommand{\paragraph}[1]{\fp {\bf #1}}

\definecolor{empcolor}{gray}{0.90}

\definecolor{notecolor}{HTML}{99CCFF}

\definecolor{warncolor}{HTML}{FF99CC}

\definecolor{donecolor}{HTML}{CC99FF}

\newcommand{\brm}[1]{\boldsymbol{\mathrm{#1}}} % bold Roman symbol
\newcommand{\lfrobsq}[1]{{\Vert#1\Vert}_{F}^{2}} %  Frobenius square norm 
\newcommand{\lnorm}[2]{{\Vert#1\Vert}_{#2}} % norm
\newcommand{\ltwo}[1]{\lnorm{#1}{2}} % l2 norm
\newcommand{\argmin}[1]{\underset{#1}{\operatorname{argmin}}\;}

\newcommand{\prox}{\operatorname{prox}}
\DeclareMathAlphabet{\mathdata}{OMS}{cmsy}{m}{n}

\def\X{\brm{X}}

\def\Y{\brm{Y}}

\def\w{\brm{w}}
\def\W{\brm{W}}

\def\I{\brm{I}}

\def\U{\brm{U}}

\def\Z{\brm{Z}}

\begin{document}

% \title{Estimating the Socioeconomic Impact of COVID-19 from Social Media}
\title{\LaTeX: Language Pattern-aware Triggering Event Detection for Adverse Experience during Pandemics}

\author{Kaiqun Fu$^1$, Yangxiao Bai$^1$, Weiwei Zhang$^2$, Deepthi Kolady$^3$\\$^1$Department of Electrical Engineering and Computer Science, South Dakota State University, USA\\$^2$School Of Psychology, Sociology And Rural Studies, South Dakota State University, USA\\$^3$Department of Agricultural Economics, Oklahoma State University, USA\\\textit{\{kaiqun.fu, bai.yangxiao, weiwei.zhang\}@sdstate.edu, deepthi.kolady@okstate.edu}}

\maketitle
\begin{abstract}
The COVID-19 pandemic has accentuated socioeconomic disparities across various
racial and ethnic groups in the United States. While previous studies have
utilized traditional survey methods like the Household Pulse Survey (HPS) to
elucidate these disparities, this paper explores the role of social media
platforms in both highlighting and addressing these challenges. Drawing from
real-time data sourced from Twitter, we analyzed language patterns related to
four major types of adverse experiences: loss of employment income (LI), food
scarcity (FS), housing insecurity (HI), and unmet needs for mental health
services (UM). We first formulate a sparsity optimization problem that
extracts low-level language features from social media data sources. Second,
we propose novel constraints on feature similarity exploiting prior knowledge
about the similarity of the language patterns among the adverse experiences.
The proposed problem is challenging to solve due to the non-convexity
objective and non-smoothness penalties. We develop an algorithm based on the
alternating direction method of multipliers (ADMM) framework to solve the
proposed formulation. Extensive experiments and comparisons to other models
on real-world social media and the detection of adverse experiences justify
the efficacy of our model. 
\end{abstract}

\section{Introduction}
\label{sec:intro}
The \textit{COVID-19} pandemic and its induced mitigation measures such as
lockdowns and social distancing had unprecedented impacts on socioeconomic
environments and resulted in increases in adverse conditions and events such
as loss of employment, reduced income, food shortage, housing insecurity,
delayed/unmet needs of healthcare in the past two years. Researchers rely
largely on population surveys and administrative records to estimate the
prevalence of a specific characteristic/outcome/condition in the general
population. These types of data are often collected in a retrospective
manner, however, lack the power to produce real-time estimates that can be
used for monitoring changes over a short period of time. In this paper, we
explore to assess the maximum utility of social media data for monitoring
temporal changes promptly and its limitations as well. We will compare the
temporal trends of adverse experiences captured in social media data to the
prevalence estimates yielded by the Household Pulse Survey (\textit{HPS})
(U.S. Census Bureau 2020). The authors will update the prevalence estimates
of four types of adverse experiences for 54 discrete time points in the past
2 years~\cite{kolady2021spatially}. 
% (See the details for the data source, methods, and results of 49 time points in Zhang and Deepthi 2022)

The rampaging {\it COVID-19} pandemic has greatly affected emotion and
everyone's daily life in the past two years. With the abundance of generated
social media data during the pandemic, more users intend to express their
emotions and opinions about social events, such as mitigation policies or the
invention of vaccines. Such dramatic social media data increase provides
great research opportunities in social media mining and natural language
processing. To further understand the various public behavior, some
researchers built a real-time tweets analyzer to get high-frequency words and
polarity over time in the United States~\cite{kabir2020coronavis}. Previous
work by Qazi~\etal~\cite{qazi2020geocov19} employs a gazetteer-based approach
to infer the geolocation of tweets. Accordingly, we deploy an
{\it ElasticSearch} search engine server to manage all the geo-tagged tweets
by time and spatial coordinates, which makes it possible to select the
appropriate study scope as needed. To mine the information and topics in
these tweets, we looked at some of the existing data mining models~\cite
{ordun2020exploratory}. 

Event detection in specific domains using social media has garnered
significant attention over the past decade~\cite
{sakaki2010earthquake,zhao2014unsupervised}. Most existing research aims to
identify events related to specific themes, such as \textit
{earthquakes}, \textit{disease outbreaks}, or \textit{civil unrest}, focusing
on events with consistent representation types. To our understanding, our
paper is the first to detect initiating events for various adverse
experiences during the \textit{COVID} era. We employ chosen keywords from
social media to capture the linguistic trends in discussions related
to \textit{COVID}. In constructing the detection model, we incorporate a
multi-task learning framework to capture the interconnectedness between
various adverse experience-related online conversations. This approach stems
from the observation that many discussions exhibit overlapping linguistic
patterns, and the language used in \textit{Twitter} posts about adverse
experiences tends to be analogous. Figure X illustrates the common linguistic
trends observed in discussions about loss of employment income (\textit
{LI}) and food scarcity (\textit{FS}).In line with the previously mentioned
specifications, we propose \textit{\underline{\textbf
{La}}nguage Pattern-aware \underline{\textbf{T}}riggering Event Detection for
Adverse \underline{\textbf{Ex}}perience}(\LaTeX) model based on multi-task
learning framework. Our main contributions are: 

$\bullet$ \textbf{Formulating a novel machine learning framework for
triggering event detection using natural language features}. Differing from
current approaches, we define the challenge of identifying initiating events
for adverse experiences during the \textit{COVID} period as a multi-task
supervised learning issue. In our suggested techniques, models for various
adverse experience-related online discussions are concurrently learned using
a shared set of linguistic features. 

$\bullet$ \textbf{Modeling similarity among various adverse experiences via
common language patterns in feature space}. Given the common linguistic
trends observed in discussions about \textit{COVID} and adverse experiences
on \textit{Twitter}, we have crafted specific constraints to capture the
similarities in language patterns across these experiences. These
resemblances in the feature domain are influenced by recurring keywords
present in online dialogues. 

$\bullet$ \textbf{Developing an efficient ADMM algorithm to learn sparse model
parameters}. The foundational optimization challenge of our suggested
multi-task model is intricate, characterized by its non-smooth, multi-convex,
and inequality-constrained nature. By incorporating auxiliary variables, we
devise an efficient ADMM-driven algorithm that breaks down the primary
challenge into multiple sub-problems. These can then be addressed using block
coordinate descent and proximal operators. 

The rest of our paper is structured as follows. related works are reviewed in
Section~\ref{sec:related}. In Section~\ref{sec:probstate}, we describe the
problem setup of our work. In Section~\ref{sec:method}, we present a detailed
discussion of our proposed \LaTeX model for detecting the occurences of the
triggering events of the adverse experiences during \textit{COVID} and
long \textit{COVID} eras, and its solution for parameter learning. In
Section~\ref{sec:experiment}, extensive experiment evaluations and
comparisons are presented. In the last section, we discuss our conclusion and
directions for future work. 

\section{Related Work}\label{sec:related}
In this section, we offer an in-depth examination of contemporary research,
differentiating between methods for detecting triggering events for adverse
experiences via social media. Specifically, we focus on local event
identification using Twitter and the application of multi-task learning
frameworks for social event detection. 

\subsection{Local Event Detection and Monitoring on Twitter}
Numerous studies in the past have leveraged social media data to detect
events. Sakaki~\etal~\cite{sakaki2010earthquake} developed a model to
determine if a recent tweet mentions an earthquake. Zhang~\etal~\cite
{zhang2015city} used taxi trace records to deduce the presence of social
events and also introduced a model to gauge the magnitude and repercussions
of these events. Santillana~\etal~\cite{santillana2015combining} investigated
the potential of machine learning in monitoring influenza activity by
harnessing data from Google searches, Twitter, and hospital records.
Gerber~\etal~\cite{gerber2014predicting} employed a logistic regression model
to predict criminal activities from a spatiotemporal standpoint. Both
Paul~\etal~\cite{paul2016social} and Parker~\cite
{parker2013framework} introduced techniques to monitor public health using
social media data. Chen~\etal~\cite{chen2016syndromic} suggested alternatives
to traditional unsupervised machine learning methods, like Latent Dirichlet
Allocation, for public health event detection, demonstrating improved flu
trend and peak predictions by aggregating regional user data over time. While
existing methodologies have proven the efficacy of social media data in
forecasting various domains, such as infectious disease spread~\cite
{chen2016syndromic,parker2013framework,santillana2015combining}, crime~\cite
{gerber2014predicting}, and disaster tracking~\cite
{sakaki2010earthquake}, there's a noticeable gap in research on adverse
experience triggering event detection via social media. To the best of our
understanding, our research is at the forefront of merging social media
analysis with multi-task learning to identify events leading to adverse
experiences during the COVID-19 and long COVID phases. Our algorithm honed
using past mitigation event data~\cite{cheng2020covid}, stands out in its
ability to promptly recognize initiating events and predict the subsequent
outcomes. 

\subsection{Multi-task Learning for Social Event Detection}
Multi-task learning (MTL) involves training models on several interconnected
tasks at once to enhance overall performance. Over recent years, numerous MTL
strategies have been introduced~\cite{zhou2011malsar}. Evgeniou~\etal~\cite
{evgeniou2004regularized} put forward a regularized MTL where the models for
all tasks are closely aligned. Task associations can also be represented by
having multiple tasks share a foundational structure, such as a unified
feature set~\cite{argyriou2006multi} or a shared subspace~\cite
{ando2005framework}. Zhao~\etal~\cite{zhao2015multi} crafted an MTL framework
tailored for forecasting tasks across related geographical locations. MTL
techniques have found applications in diverse fields, from computer vision to
biomedical informatics. As far as we're aware, our method is the pioneering
supervised learning framework that captures the language interplay under the
context of triggering events of adverse experiences during COVID-19 and long
COVID phases through the lens of the multi-task learning approach. 

\section{Problem Setup}
\label{sec:probstate}
Given a collection of tweets $\mathcal{T}$, which is collected along a
continuous time series, we first filter it using a set of COVID-19 related
keywords and keywords mentioned in the {\it CORONANet} project~\cite{unknown}. This
produces the target tweet subcollection $\mathcal{T}^{+}$. Then based on
which COVID-19-related events are referred to in each tweet, $\mathcal{T}^
{+}$ is grouped into $\{\mathcal{T}_{C}^{+}\}^{C\in\Phi}$, where $\Phi=
\{\texttt{LI},\texttt{FS},\texttt{HI},\texttt{UM}\}$ represents four types for
adverse experiences:{\it loss of employment income} (\texttt{LI}), {\it food
scarcity} (\texttt{FS}), {\it housing insecurity} (\texttt{HI}), and
{\it unmet needs for mental health services} (\texttt{UM}). 

In this work, we address two major questions: 1) given a type of adverse
experience $c$, a time slot $t$, and the collection of corresponding tweets
$\mathcal{T}_{c,t}^{+}$, is there a corresponding event that triggers such
adverse experience $c$ during time period $t$? To answer this question, we
will cast it as a supervised learning problem using the multi-task learning
framework; and 2) given a set of classified Twitter set $\mathcal{T}_{t}^
{+}$ in the time period $t$, what is sentiment score for that time period? To
answer this question, we will also format it as a supervised learning problem
but applying the recurrent neural network models. 

Under the assumption that the COVID-19-related adverse experiences can be
captured by complaints and negative discussion in Twitter space, we adopt a
dictionary $\mathcal{F}$ of features trained specifically for Twitter~\cite
{mohammad2013nrc}. For each subcollection $\mathcal{T}_{c,t}^{+}$, we
generate a corresponding matrix $\X_t^c$ by counting the frequencies of
semantic features in $\mathcal{F}$. Now, our problem can be formulated by
performing the mapping:
\begin{equation}
F_c(\X_t^c)\rightarrow\Y_t^c
\end{equation}
where $\Y_t^c\in\{-1, 1\}$ are labels that denote the occurrence of a
COVID-19-related triggering event, and $F_c$ is the model for the adverse
experience type $c$. 
% (As mentioned previously, there are four types from the \warn{{\it HPS} Survey}.)

A traditional way to solve this problem is to learn the model for each adverse
experience separately. However, the performance of each model may be affected
by ignoring the relatedness among different types. In our approach, this
relatedness is expressed as the semantic similarities among complaints about
different COVID-19-related adverse experiences. We consider that two factors
contribute to this semantic similarity in Twitter space. \textbf{(1) Topical
relatedness of adverse experiences}: the adverse experiences are topically
related together (\eg, some COVID-19 events may trigger multiple negative
emotions from different adverse experiences). As a result, the triggering
events that affect multiple experiences may provoke similar
complaints. \textbf{(2) Common complaint vocabulary targetting COVID-19}: We
assume that the words used by Twitter users to complain about COVID-19, in
general, will be similar across all adverse experiences. To model semantic
similarity caused by these two factors, we propose a multitask learning-based
COVID-19 triggering events detection model. 

\section{Proposed Method}
\label{sec:method}
Considering that we want to predict if there is a triggering event for a
specific type of adverse experience resulting from COVID-19, given a
subcollection of tweets $\mathcal{T}_{c,t}^{+}$ which includes the specific
adverse experience $c$ during time slot $t$, our problem fits well into the
scope of a classification or regression problem. For instance, learning the
function $F_c$ can be modeled as a logistic regression problem and the model
parameters $\w$ can be learned by solving the following optimization
problem:
\begin{equation}
\argmin{\w}\mathcal{L}_{c}=\sum_{t=1}^{m_c}log(1+exp\{\Y_t^c(\X_t^c\w)\})
\label{eq:opt_prob}
\end{equation}
where $m_c$ is the total number of data points in $\mathcal{T}_{c,t}^
{+}$. However, as stated in Section~\ref{sec:probstate}, if $\w$ for each
experience is learned separately, these models will fail to reflect the
semantic similarity among the experiences in feature space. To solve these
challenges, we case the original problem into a multi-task learning
framework: 
\begin{equation}
\argmin{\w}\mathcal{L}=\sum_{c=1}^{|\Phi|}\sum_{t=1}^{m_c}log(1+exp\{-\Y_t^c(\X_t^c\W^c)\})
\label{eq:multi_prob}
\end{equation}
where each column of $\W$, referred to as $\W^c$, denotes the model parameters
of $F_c$. In this way, we can further model the relatedness among the adverse
experiments with parameter matrix $\W$. 

\subsection{Modeling Language Pattern Similarity Between Adverse Experiences}
\label{subsec:between_experiences}
While studying the COVID-19 resulting adverse experiences, distinct
experiences are often consequentially related. That is, two or more adverse
experiences may be caused by the same type of COVID-19-related events. For
instance, the experience type \textit{loss of employment income} (\texttt
{LI}) and \textit{housing insecurity} (\texttt{HI}) may have a similar
language pattern on social media platforms when discussed. This means that if
a \textit{loss of employment income} triggering event occurs, there is a
good chance that such an event will also conduct some impact on housing
insecurity experiences. The correlation between adverse experiences results
in semantic similarity in Twitter space and, therefore, a similar
distribution of tweets complaining or discussing adverse experiences.
Thus, our model should be encouraged to capture this form of semantic
relatedness in Twitter space. Mathematically, we place constraints on
parameters among different tasks: 
\begin{equation}
\begin{split}
\argmin{\W}&\sum_{c=1}^{|\Phi|}\sum_{t=1}^{m_c}\mathrm{log}(1+\mathrm{exp}\{-\Y_t^c(\X_t^c\W^c)\})\\
\mathrm{s.t.}\text{\ } &||\W^1-\W^2||_2^2\leq\eta_1,||\W^1-\W^3||_2^2\leq\eta_2\\
&||\W^2-\W^3||_2^2\leq\eta_3,||\W^2-\W^4||_2^2\leq\eta_4\\
&\eta_1\geq0,\eta_2\geq0,\eta_3\geq0,\eta_4\geq0
\label{eq:multitask_prob1}
\end{split}
\end{equation}

\subsection{Modeling Common Word Features in Adverse Experiences in Feature Space}
\label{subsec:common_feat}
In addition to similarities between the adverse experiences, we also consider
a hidden pattern in the usage of complaint words over time and for COVID-19
in general. These were posted at different timestamps and, although they
complain about \textit{loss of employment income} (\texttt{LI}) and \textit
{housing insecurity} (\texttt{HI}) respectively, they share a common
complaint word. This observation leads us to believe that although we have
adopted a large dictionary of semantic keywords, it is possible that only a
small subset of them, contribute to the detection of general adverse
experiences. This means that the learned parameters matrix W should be sparse
and have nonzero values for only the most important features. Thus, the
proposed model should be encouraged to capture hidden patterns among
complaints and to maintain sparsity in feature space. Mathematically, this
consideration inspires us to use the $\ell_{2,1}$ norm~\cite
{argyriou2008convex} to perform joint feature selection: 
\begin{equation}
\begin{split}
\argmin{\W}&\sum_{c=1}^{|\Phi|}\sum_{t=1}^{m_c}\mathrm{log}(1+\mathrm{exp}\{-\Y_t^c(\X_t^c\W^c)\})\\
\mathrm{s.t.}\text{\ } &||\W||_{2,1}\leq\eta_5,\eta_5\geq0
\label{eq:multitask_prob2}
\end{split}
\end{equation}

Combining Model~(\ref{eq:multitask_prob1}) and Model~(\ref{eq:multitask_prob2}) together, we get our proposed model: 
\begin{equation}
\begin{split}
\argmin{\W}&\sum_{c=1}^{|\Phi|}\sum_{t=1}^{m_c}\mathrm{log}(1+\mathrm{exp}\{-\Y_t^c(\X_t^c\W^c)\})\\
\mathrm{s.t.}\text{\ } &||\W^1-\W^2||_2^2\leq\eta_1,||\W^1-\W^3||_2^2\leq\eta_2\\
&||\W^2-\W^3||_2^2\leq\eta_3,||\W^2-\W^4||_2^2\leq\eta_4\\
&||\W||_{2,1}\leq\eta_5\\
&\eta_1\geq0,\eta_2\geq0,\eta_3\geq0,\eta_4\geq0,\eta_5\geq0
\label{eq:multitask}
\end{split}
\end{equation}
We can then move the constraints to the objective function, we will obtain an equivalent regularized problem, which is easier to solve: 
\begin{equation}
\begin{split}
\argmin{\W}&\sum_{c=1}^{|\Phi|}\sum_{t=1}^{m_c}\mathrm{log}(1+\mathrm{exp}\{-\Y_t^c(\X_t^c\W^c)\})+\lambda_5||\W||_{2,1}\\
&+\lambda_1||\W^1-\W^2||_2^2+\lambda_2||\W^1-\W^3||_2^2\\
&+\lambda_3||\W^2-\W^3||_2^2+\lambda_4||\W^2-\W^4||_2^2
\label{eq:multitask_constrain}
\end{split}
\end{equation}
where $\lambda_1$, $\lambda_2$, $\lambda_3$, $\lambda_4$, and $\lambda_5$ are trade-off penalties balancing the value of the loss function and the regularizers. 

\subsection{Parameter Learning}
\label{subsec:para_learn}
The objective function in Equation~\ref{eq:multitask_constrain} is
multi-convex and the regularizer $\ell_{2,1}$ is non-smooth. This increases
the difficulty of solving this problem. A traditional way to solve this kind
of problem is to use proximal gradient descent. But this approach is slow to
converge. Recently, the alternating direction method of multipliers
(ADMM)~\cite{boyd2011distributed} has become popular as an efficient
algorithm framework that decouples the original problem into smaller and
easier-to-handle subproblems. Here we propose an ADMM-based Algorithm~\ref
{alg:admm} which can optimize the proposed models efficiently. In particular,
primal variables are updated on Line 4, dual variables on Line 5, and
Lagrange multipliers on Line 6. Line 7 calculates both primal and dual
residuals.

\begin{algorithm}[htpb]
    \KwIn{$\X, \Y$}
    \KwOut{$\W$}
    Initialize $\W^{(0)}$, $\U_{\W}^{(0)}$, $\mathbf{\Lambda}_{1}^{(0)}$\;
    Initialize $\rho = 1$, $\epsilon^{p} > 0, \epsilon^{d} > 0$, $\mathrm{MAX\_ITER}$\;
    \For{$k = 1 : \mathrm{MAX\_ITER}$}
    {%
        %\tcp{Update primal variables.}
        Update $\W^{(k)}$ with BCD using Equations~\ref{equa:grad_s}\; 
        %\tcp{Update dual variables.}
        Update $\U_{\W}^{(k)}$ with Equations~\ref{equa:update_dual}\;
        %\tcp{Update Lagrangian multiplier.}
        Update $\mathbf{\Lambda}_{1}^{(k)}$ with Equations~\ref{equa:update_lm}\;
        Compute $p$ and $d$ by Equations~\ref{equa:update_residuals}\;
        \If{$p < \epsilon^{p}$ and $d < \epsilon^{d}$}
        {%
            break\;
        }
    }
    \caption{An ADMM-based solver}\label{alg:admm}
\end{algorithm}

\subsection{Augmented Lagrangian Scheme}%
\label{subsec:augmented_lagrangian_scheme}

First, we introduce an auxiliary variable $\U_{\W} = \W$
into the original problem~\ref{eq:multitask_constrain} and obtain the following equivalent
problem:
\begin{equation}\label{equa:auxiliary_variables}
    \begin{aligned}
        \argmin{\Theta} & \sum_{c=1}^{|\Phi|}\sum_{t=1}^{m_c}\mathrm{log}(1+\mathrm{exp}\{-\Y_t^c(\X_t^c\W^c)\})\\
		&+\lambda_1||\W^1-\W^2||_2^2+\lambda_2||\W^1-\W^3||_2^2\\
		&+\lambda_3||\W^2-\W^3||_2^2+\lambda_4||\W^2-\W^4||_2^2\\
		&+\lambda_5||\U_{\W}||_{2,1}\\
        \mathrm{s.t.}\text{\ } & \U_{\W} = \W, 
    \end{aligned}
\end{equation}
where $\Theta = \{\W, \U_{\W}\}$ is the set of variables to be optimized. Then
we transform the above problem into its augmented Lagrangian form as
follows:
\begin{equation}\label{equa:lagrange}
    \begin{aligned}
        & \sum_{c=1}^{|\Phi|}\sum_{t=1}^{m_c}\mathrm{log}(1+\mathrm{exp}\{-\Y_t^c(\X_t^c\W^c)\})\\
		&+\lambda_1||\W^1-\W^2||_2^2+\lambda_2||\W^1-\W^3||_2^2\\
		&+\lambda_3||\W^2-\W^3||_2^2+\lambda_4||\W^2-\W^4||_2^2\\
		&+\lambda_5||\U_{\W}||_{2,1}+\langle\mathbf{\Lambda}_1, \W-\U_{\W}\rangle\\
        &+\frac{\rho}{2}\lfrobsq{\W-\U_{\W}}
    \end{aligned} 
\end{equation}
where $\mathbf{\Lambda}_1$ is the Lagrangian multipliers. With this step, we
decouple the original problem into two easier-to-handle problems in which
seven variables $\W$, $\U_{\W}$, and $\mathbf{\Lambda}_1$ will be optimized
individually. 

\subsection{Parameter Optimization}
\label{subsec:parameters_optimization}
The Lagrangian form in Equation~\ref{equa:lagrange} is separated based on the
primal variables and the dual variables, where the problem of solving the
primal variable $\W$ is smooth and convex. 

\begin{table*}[htpb!]
    \large
    \centering
\begin{tabular}{lcc|cc|cc|cc}
\hline
\multicolumn{1}{c}{\multirow{2}{*}{Method}} & \multicolumn{2}{c|}{\texttt{LI}}           & \multicolumn{2}{c|}{\texttt{FS}}           & \multicolumn{2}{c|}{\texttt{HI}}           & \multicolumn{2}{c}{\texttt{UM}}            \\ \cline{2-9} 
\multicolumn{1}{c}{}                        & Precision       & F1              & Precision       & F1              & Precision       & F1              & Precision       & F1              \\ \hline
\multicolumn{1}{l|}{LASSO}                  & 0.2941          & 0.4545          & 0.5556          & 0.7143          & 0.6286          & 0.7719          & 0.5946          & 0.7458          \\
\multicolumn{1}{l|}{Ridge}                  & 0.6500          & 0.5333          & 0.7043          & 0.8242          & 0.4909          & 0.5481          & 0.5250          & 0.5481          \\
\multicolumn{1}{l|}{SVC}                    & 0.5946          & 0.7458          & 0.5556          & 0.7143          & 0.2941          & 0.4545          & 0.6286          & 0.7719          \\
\multicolumn{1}{l|}{\LaTeX}              & \textbf{0.6970} & \textbf{0.7797} & \textbf{0.7297} & \textbf{0.8437} & \textbf{0.8611} & \textbf{0.9118} & \textbf{0.7027} & \textbf{0.8254} \\ \hline
\end{tabular}
\caption{COVID-19-Triggered Adverse Experiences Comparisons with Different Topics ({\it loss of employment income} (\texttt{LI}), {\it food
scarcity} (\texttt{FS}), {\it housing insecurity} (\texttt{HI}), and
{\it unmet needs for mental health services} (\texttt{UM}))}\label{tab:eval}
\end{table*}

\subsubsection{Update $\W$}
\label{subsub:update_W}
We define Equation~\ref{equa:lagrange} as objective function $\mathcal{Q}$
which is multi-convex. In particular, $\mathcal{Q}$ of $\W^{c}$ is convex
where all other $\W^{c' \neq c}$ are fixed. This kind of problem can be
decoupled into subproblems using block coordinate descent
(BCD)~\cite{xu2013block}, in which each $\W^{c}$ is updated by solving the
following sub-optimization problems:
\begin{equation}
    \W^{c} \leftarrow \argmin{\W^{c}} \mathcal{Q}. 
\end{equation}
$\mathcal{Q}$ is smooth and convex for each $\W_{r}$ and can be solved by
gradient descent as follows: 
\begin{equation}\label{equa:grad_s}
    \frac{\partial \mathcal{Q}}{\partial \W_{i}} =
        \mathcal{P}(i) + 2\sum_{ij}\mathdata{M}_{ij}\cdot\lambda_k(\W_{i} - \W_{j})
\end{equation}
where according to the BCD algorithm, the $\partial\mathcal{Q}_{\W}/\partial
\W_{i}$ is calculated in sequence, from $i=1$ to $k$. And the
$\mathcal{P}(c)$ is defined as follows:
\[
    \begin{aligned}
        \mathcal{P}(c) = &\X_c^T(-\Y^c\circ(\I-\I/(\I+\mathrm{exp}\{\Z^c\})) \\ 
        &+\mathbf{\Lambda}_{1}^{c}+\rho(\W_{c}-\U_{\W}^{c})
    \end{aligned}
\]
where $\mathbf{\Lambda}_{1}^{c}$ and $\U_{\W}^{c}$ are the $c$-th columns of
the corresponding Lagrangian multiplier and dual variable; $\circ$ is the
element-wise product (Hadamard product); $\I$ is a $m_c$-dimensional identity
vector, and $\Z^c$ is defined as $\Z^c=-\Y^c\circ(\X^c\W^c)$. 

\subsubsection{Update Dual Variables}
\label{subsub:update_dual}
Now that primal variable $\W$ is taken care of, the dual variable $\U_
{\W}$ are updated as follows: 
\begin{equation}\label{equa:update_dual}
    \U_{\W}^{+} \leftarrow \prox_{f_{1}, 1/\rho}(\mathbf{\Lambda}_{1} + \W)
\end{equation}
where $f_{1}$ is the non-smooth function $\lambda_{\W}\lnorm{\U_{\W}}{2, 1}$
The proximal operator can be solved efficiently using proximal
operators~\cite{parikh2014proximal}.

Next, the Lagrangian multiplier $\mathbf{\Lambda}_{1}$ is updated as follows:
\begin{equation}\label{equa:update_lm}
	\mathbf{\Lambda}_{1}^{+} \leftarrow \mathbf{\Lambda}_{1} + \rho(\W^{+} - \U_{\W}^{+})
\end{equation}
Finally, primal and dual residuals are calculated with: 
\begin{equation}\label{equa:update_residuals}
    p = \ltwo{\W^{+} - \U_{\W}^{+}}, 
    d = \rho\left(\ltwo{\U_{\W}^{+} - \U_{\W}}\right).
\end{equation}
where $p$ is primal residual, and $d$ is dual residual. 

\section{Experiment}\label{sec:experiment}
In this section, we present the experiment environment, dataset introduction,
evaluation metrics and comparison methods, extensive experimental analysis on
predictive results, and discussions on the learner features.

\subsection{Experiment Setup}
\label{ssub:experiment_setup}
\subsubsection{Experiment Environment}
\label{subsub:experiment_environment}
We conducted our experiments on a machine with an Intel Core i9 CPU with 3.70
GHz, the computational power of this CPU is 4.13 Gflops per core. Time
requirements should be an important factor for real-world triggering event
detection for \textit{COVID-19}-related adverse experiences. The most
time-consuming process of our proposed model is at the training stage. The
training stage learns the parameters for temporal features $\W$. A matrix
multiplication $\X\W$ will generate the prediction rapidly. Our prediction
for a single data point is generated in less than $0.001$ seconds in the
validation and testing stages. 

\subsubsection{Feature Settings and Dataset}%
\label{subsub:dataset_and_feature}
The semantic similarities among complaints about different
\textit{COVID-19}-related adverse experiences (\texttt{LI}, \texttt{FS}, \texttt
{HI}, and \texttt{UM}) are determined by the cosine similarities of the
keywords of the collected tweets of each topic. We found the following task
pairs with high cosine similarities: \texttt{LI}-\texttt{FS}, \texttt
{LI}-\texttt{HI}, \texttt{FS}-\texttt{HI}, and \texttt{FS}-\texttt
{UM}. Consequently, we model the problem with the following $\ell_
{2}$ regularizers: $||\W^1-\W^2||_2^2\leq\eta_1$,
$||\W^1-\W^3||_2^2\leq\eta_2$, $||\W^2-\W^3||_2^2\leq\eta_3$, and
$||\W^2-\W^4||_2^2\leq\eta_4$. 

We utilized the \textit{CORONAVIRUS (COVID-19) GEO-TAGGED TWEETS DATASET}
obtained from \textit{IEEE DataPort}~\cite{lamsal2021design} for our
research. This dataset consists of English tweets pertaining to the epidemic
worldwide, spanning from 2020 to 2022. Since direct access to Twitter content
is restricted due to their dissemination policy, we employed the \textit
{Twarc} Python package to retrieve complete information, including tweet
content, UTC time, and geographical location, using the tweet IDs. A total of
469,414 tweets were imported into an \textit{Elastic} server. To establish a
filtering framework, we consulted the \textit{COVID-19} Government Response
Database~\cite{cheng2020covid} and extracted keywords from policy
announcements. Subsequently, we developed a comprehensive filtering function
to generate appropriate study samples.

\subsection{Comparison Methods and Evaluation Metrics}%
\label{sub:comparison_methods}
To justify our model's performance in predicting the occurrences of
COVID-19-triggered adverse experiences, three comparison methods are
considered in our experiment: $\ell_2$ regulized linear regression (\textit
{Ridge regression}), $\ell_1$ regulized linear regression (\textit
{LASSO}), and support vector regression (\textit{SVC}). 

$\bullet$ \textbf{$\ell_2$ Regulized Linear Classification (Ridge)}~\cite
{peeta2000providing}. Ridge classification is an extension of linear
regression. It's a linear classification model regulized on $\ell_2$ norm.
The $\lambda$ parameter is a scalar that controls the model complexity; the
smaller $\lambda$ is, the more complex the model will be. In our
implementation, $\lambda$ is searched from $\{10, 100\}$. 

$\bullet$ \textbf{$\ell_1$ Regulized Linear Classification (LASSO)}~\cite
{ramakrishnan2014beating,tibshirani1996regression}. This is a classic way to
conduct cost-efficient regressions by enforcing the sparsity of the selected
features. It has been proven to be effective in the field of event
detection~\cite{ramakrishnan2014beating}. It includes a parameter $\lambda$
that trades off the regularization term; typically, the larger this parameter
is, the fewer the selected features will be. In our experiment, $\lambda$ is
searched from $\{1, 10, 100\}$. The feature configurations applied by this
model are the same as the ridge classification model. 

$\bullet$ \textbf{Support Vector Classification (SVC)}~\cite
{tibshirani1996regression}. Support vector classification provides solutions
for both linear and non-linear problems. In our experiment implementation, we
utilize non-linear kernel functions (RBF kernel) to find the optimal solution
for the adverse experience prediction problem. The model parameters are
selected with $c=1$ and $\epsilon=0.1$. This model considers similar temporal
features with ridge classification and LASSO methods, no multi-task features
for inter-topic semantic similarity are considered. 

To quantify and validate the performance of our model, we adopt Precision and
F-measure ($F_1$). These metrics are widely utilized in the field of event
prediction studies~\cite
{li2018overview,khattak2016modeling,park2016interpretation,zou2016application},
it reflects the predictive performance of the proposed model. 

\subsection{Adverse Experiences Triggering Event Detection}%
\label{sub:detection_results}
\subsubsection{Multitask Learning Performance Analysis on Triggering Events Detection}
\label{subsub:titan_connect}
Table~\ref{tab:eval} presents a comprehensive comparison between our proposed
method and other competing approaches for detecting triggering events related
to adverse experiences. The experimental results provide substantial evidence
to support our utilization of a multi-task learning framework for predicting
the occurrences of these triggering events. Overall, the multi-task learning
method demonstrates superior performance in terms of Precision and F1 score
compared to single-task models such as LR, SVC, and LASSO. This outcome
highlights the significance of incorporating semantic similarities among
COVID-19-related adverse experiences, as it significantly enhances the
detection accuracy of triggering events. Notably, the multi-task learning
method outperforms the LASSO model in both Precision and F1 scores. These
findings emphasize that, for effective triggering event detection, relying
solely on $\ell_1$ regularizers is insufficient, and it is crucial to
incorporate detailed semantic similarity modeling in the context of
COVID-19-related topic discussions.

\subsubsection{Performance Analysis between Training Tasks}
\label{subsub:task_perform}
The results in Table~\ref{tab:eval} show that the model performance for
triggering event detection is not the same across different COVID-19-related
topic discussions. For instance, the prediction performances of all the
comparison methods on the topic \textit{Loss of Income} (\texttt{LI}) are
relatively lower than other topics. This is because the topic \textit{Loss of
Income} has fewer corresponding triggering events than the rest of the
topics, and the constraint of semantic distances for \textit{Loss of Income}
only shares limited similarities with the other columns of the feature matrix
$\W$. In contrast, our model for the topics \textit{food scarcity}
and \textit{housing insecurity} outperforms the comparison methods, because
these tasks have more triggering events recorded. 

\section{Conclusion}
\label{sec:conclusion}
Our findings indicate that social media platforms not only mirror the adverse
experiences captured by traditional surveys but also offer a more immediate
and granular view of these challenges. Furthermore, the real-time nature of
social media platforms provides an opportunity for timely interventions, from
connecting affected individuals to resources to influencing policy decisions.
As the world becomes increasingly digital, understanding the potential of
these platforms to both reflect and address societal challenges is crucial
for a holistic and timely response to future crises. Our proposed approach
successfully extracted the focus topics related to the epidemic. It provided
a very intuitive way to show each topic's potential connections and its
geographic time information. However, the accuracy of the results is often
curbed due to the large amount of noise present in Twitter text, as well as
the limitation of text length. In future work, we optimize the model in two
ways: 1) Apply text summarization methods to conclude the topics and enhance
the interpretability of the results; and 2) Use the attention mechanism to
reduce the impact of noise on text analytics.

\bibliographystyle{abbrv}
\bibliography{bigdata_wksp_2023.bib}

\end{document}